\documentclass[10pt,twocolumn,letterpaper]{article}

\usepackage{iccv}
\usepackage{times}
\usepackage{epsfig}
\usepackage{graphicx}
\usepackage{amsmath}
\usepackage{amssymb}
\usepackage{makecell}
\usepackage[accsupp]{axessibility}  

\usepackage[pagebackref,breaklinks,colorlinks]{hyperref}

\newcommand{\bs}[1]{\boldsymbol{#1}}

\iccvfinalcopy 

\def\iccvPaperID{****} 

\ificcvfinal\fi
\begin{document}

\title{Linear-Covariance Loss for End-to-End Learning of 6D Pose Estimation}

\author{Fulin Liu\\
Beihang University\\
{\tt\small liufulin@buaa.edu.cn}
\and
Yinlin Hu\\
MagicLeap\\
{\tt\small yhu@magicleap.com}
\and
Mathieu Salzmann\\
EPFL, ClearSpace\\
{\tt\small mathieu.salzmann@epfl.ch}
}

\maketitle
\ificcvfinal\thispagestyle{empty}\fi

\begin{abstract}
Most modern image-based 6D object pose estimation methods learn to predict 2D-3D correspondences, from which the pose can be obtained using a PnP solver. Because of the non-differentiable nature of common PnP solvers, these methods are supervised via the individual correspondences. To address this, several methods have designed differentiable PnP strategies, thus imposing supervision on the pose obtained after the PnP step. Here, we argue that this conflicts with the averaging nature of the PnP problem, leading to gradients that may encourage the network to degrade the accuracy of individual correspondences. To address this, we derive a loss function that exploits the ground truth pose before solving the PnP problem. Specifically, we linearize the PnP solver around the ground-truth pose and compute the covariance of the resulting pose distribution. We then define our loss based on the diagonal covariance elements, which entails considering the final pose estimate yet not suffering from the PnP averaging issue. Our experiments show that our loss consistently improves the pose estimation accuracy for both dense and sparse correspondence based methods, achieving state-of-the-art results on both Linemod-Occluded and YCB-Video.

\end{abstract}

\section{Introduction}
\label{sec:intro}

Estimating the 6D pose of 3D objects from monocular images is a core computer vision task, with many real world applications, such as robotics manipulation~\cite{zhu2014single,Zuo_2019_CVPR}, autonomous driving~\cite{Chen_2017_CVPR,Wu_2019_CVPR_Workshops} and augmented reality~\cite{crivellaro2017robust,marchand2015pose}.
Although this task can be facilitated by the use of RGBD input,  depth sensors are not ubiquitous, and thus 6D object pose estimation from RGB images remains an active research area. 

With the development of deep neural networks (DNNs), early methods~\cite{billings2019silhonet, do2018deep6dpose, Kehl_2017_ICCV, xiang2018posecnn} formulated pose estimation as a regression problem, directly mapping the input image to the 6D object pose.
More recently, most works~\cite{Chen_2022_WACV, Jawahar2018ipose, Hu_2019_CVPR,Merrill_2022_CVPR, Oberweger_2018_ECCV,Park_2019_ICCV,pavlakos2017,  Peng_2019_CVPR,Rad_2017_ICCV, Su_2022_CVPR,Tekin_2018_CVPR} draw inspiration from geometry and seek to predict 2D-3D correspondences, from which the 6D pose can be obtained by solving the Perspective-n-Points (PnP) problem. 
While effective, these methods supervise the training process with the individual correspondences, and not with the ground-truth pose itself, as standard PnP solvers are not differentiable.

\begin{figure}[t]
\begin{center}
\includegraphics[width=1\linewidth,trim=4 36 4 -2]{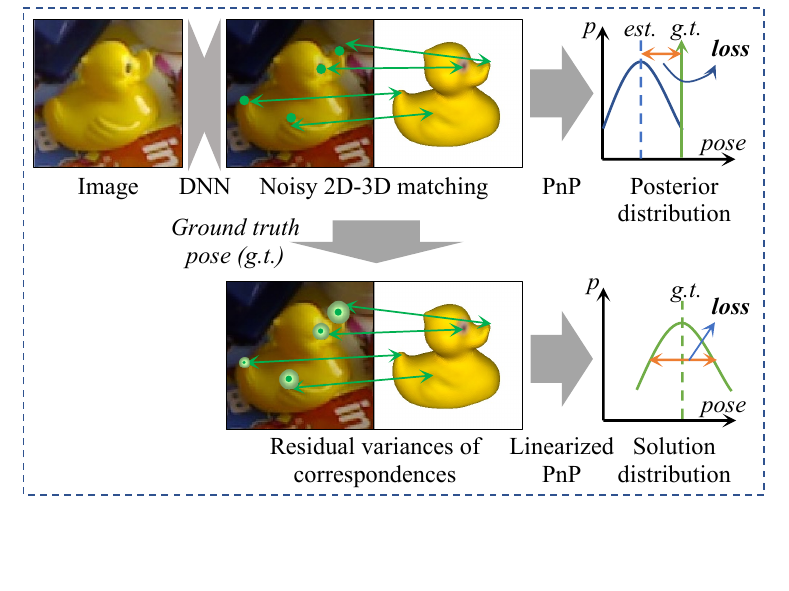}
\end{center}
   \caption{\textbf{Difference between other differentiable PnP losses and our proposed loss.} Other methods (first row) first solve for the object pose from noisy correspondences, and then compute the loss based on the resulting pose and the ground truth. By contrast, we (second row) utilize the ground-truth pose to estimate the residual variance of the correspondences and compute the covariance of the pose distribution. We define our loss based on the diagonal  elements of this covariance.}
\label{fig:full}
\end{figure}

To enable end-to-end training, several attempts have been made to incorporate the PnP solver as a differentiable network layer~\cite{Brachmann_2017_CVPR,Brachmann_2018_CVPR,Chen_2020_CVPR}.
While these methods make it possible to employ pose-driven loss functions to train the DNN, they only leverage the optimal pose as supervision, thus not imposing constraints on other pose candidates.
In~\cite{Chen_2022_CVPR}, this was addressed by deriving a loss function based on the posterior pose distribution, encouraging a larger posterior for the ground truth and smaller posteriors for the other poses.

Nevertheless, to the best of our knowledge, all of these differentiable PnP layers have a common property: They first solve the PnP problem to obtain either the pose~\cite{Brachmann_2017_CVPR,Brachmann_2018_CVPR,Chen_2020_CVPR} or the posterior pose distribution~\cite{Chen_2022_CVPR}, and then compute the error to be backpropagated based on a dedicated loss function and the ground-truth pose. That is, they introduce the ground-truth information only \emph{after} the pose has been computed. While this may seem a natural strategy when incorporating a differentiable layer, we argue that this conflicts with the averaging nature of the PnP problem, which aggregates multiple noisy measurements into a single estimate.

\begin{figure}[t]
\begin{center}
\includegraphics[width=1\linewidth,trim=4 127 4 0]{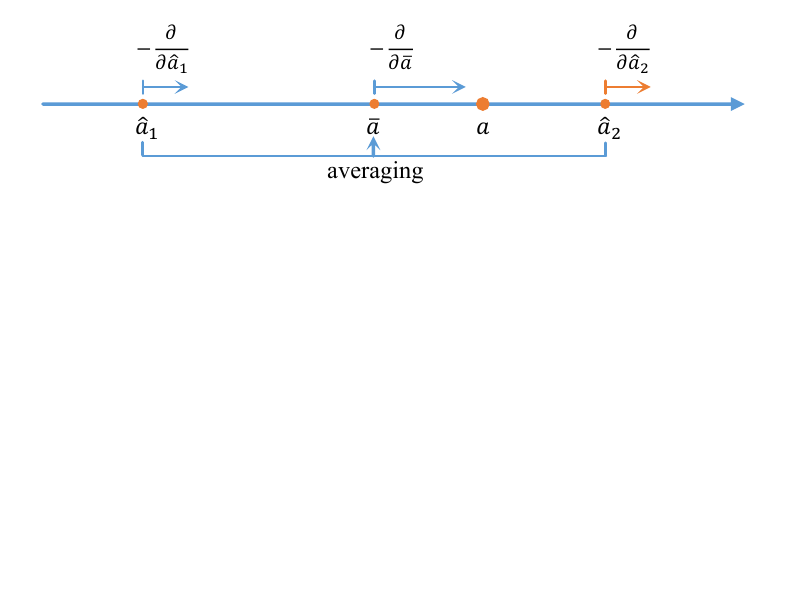}
\end{center}
   \caption{\textbf{Gradients after averaging.} In this toy example, the gradient will drive $\hat{a}_2$ away from the true $a$ value, although $\bar{a}$ is driven closer to $a$.}
\label{fig:toy}
\end{figure}
To illustrate this, let us consider a simpler problem with two noisy measurements, $\hat{a}_1$ and $\hat{a}_2$, of a value $a$, and seek to minimize the distance between the average value $\bar{a}=(\hat{a}_1+\hat{a}_2)/2$ and $a$, i.e., $|\bar{a}-a|$. As $\hat{a}_1$ and $\hat{a}_2$ contribute to the loss in the same manner, they are updated with the same gradient, irrespectively of their individual error w.r.t. $a$. For example, as depicted in Fig.~\ref{fig:toy}, if $\hat{a}_1$ and $\hat{a}_2$ are on either side of $a$, the gradient will drive one of these estimates further from the true $a$ value, i.e., degrade the individual prediction quality of this estimate.

In the case of PnP, 4 2D-3D correspondences are in general sufficient to obtain a unique pose~\cite{lepetit2009ep}. If more than 4 correspondences are provided, the PnP solver performs a form of ``averaging'', and thus may yield a similar effect as in the previous toy example: It may encourage a decrease in accuracy of some of the correspondences, thus essentially providing confusing signal in the network training process. While this can be alleviated by using the pose-driven loss in conjunction with correspondence-level supervision, this strategy circumvents the problem instead of addressing it.

By contrast, in this paper, we introduce an approach to explicitly tackle the gradient issue arising from the averaging process of the PnP problem. To this end, we leverage the covariance of the pose distribution computed by exploiting the ground-truth pose \emph{before} solving for the pose, as illustrated by Fig.~\ref{fig:full}. Given noisy 2D-3D correspondences as input to the PnP solver, we consider the distribution of PnP-computed poses around the ground-truth one. We then approximate the covariance of this distribution by linearizing the PnP solver around the ground truth. 
This lets us design a loss function that minimizes the diagonal covariance elements, which entails minimizing the 2D-3D correspondence residuals while nonetheless considering the pose estimated via PnP.
Our formalism also applies to the weighted PnP scenario, allowing us to compute weights for the individual correspondences so as to emphasize the ones that benefit the pose estimate.

Our experiments on several datasets and using both sparse and dense correspondence based methods evidence the benefits of our approach. In particular, applying our loss to the state-of-the-art ZebraPose~\cite{Su_2022_CVPR}, lets us achieve state-of-the-art performance on both the LM-O and YCB-V datasets. Our code is available at \url{https://github.com/fulliu/lc}.

\section{Related Work}
In this section, we focus on the geometry-driven methods, which are the most relevant to our core contributions. Geometry-driven methods solve the pose estimation task by first extracting 2D-3D correspondences from RGB images, and then computing the 6D pose using a PnP solver. They can be roughly categorized into sparse keypoints-based methods and dense prediction ones.

\textbf{Sparse Correspondences based Estimation.}\quad
Sparse keypoints-based methods predict the 2D projected locations of predefined 3D keypoints. Standard 3D keypoint choices include the object centroid, the corners of the object bounding box, semantic keypoints, or keypoints sampled from the object model, e.g., via the farthest point sampling algorithm~\cite{Peng_2019_CVPR}. Specifically, 
BB8~\cite{Rad_2017_ICCV} predicts the image locations of the 8 corners of the object bounding box; Bugra \etal~\cite{Tekin_2018_CVPR} add the centroid of the object 3D model to the correspondence list; Pavlakos \etal~\cite{pavlakos2017} use heatmaps to predict the 2D locations of semantic keypoints. A voting scheme is also utilized to aggregate multiple predictions for the same 2D point so as to improve robustness to partial occlusions.
Oberweger \etal~\cite{Oberweger_2018_ECCV} combine different heatmap predictions generated from different image patches; Hu \etal~\cite{Hu_2019_CVPR} aggregate the 2D keypoint predictions from all pixels belonging to the given target; Similarly, PVNet \cite{Peng_2019_CVPR} regresses the vector-field pointing from each object pixel to the 2D locations.
Chen \etal~\cite{Chen_2022_WACV} combine heatmap-based keypoint predictions at different scales and use occlude-and-blackout data augmentation to boost occlusion robustness. All of these methods use a traditional PnP solver as second stage, with the exception of Hu \etal~\cite{Hu_2020_CVPR} that replaces the PnP solver with an MLP, enabling end-to-end training of the whole pipeline.

\textbf{Dense Correspondences based Estimation.}\quad
Methods based on dense correspondences predict pixel-wise 3D object coordinates inside the object instance mask. iPose~\cite{Jawahar2018ipose} achieves this by first cropping the image patch containing the target based on the instance mask; Pix2Pose \cite{Park_2019_ICCV} uses the 2D image bounding box from a detector to crop the object, and then regress both the 3D coordinates and their errors, using generative adversarial training for occlusion robustness.
Similarly to the sparse methods, the PnP solver can be replaced by an MLP. CDPN \cite{Li_2019_ICCV} uses an MLP to regress the object translation but a PnP solver to compute the rotation.
When an MLP is used, the correspondences are not restricted to 2D-3D points. GDR-net \cite{Wang_2021_CVPR} additionally feeds surface region attention maps into an MLP-based Patch-PnP module. Based on GDR-Net, SO-Pose \cite{Di_2021_ICCV} exploits a self-occlusion map, replacing the surface region attention maps as another input modality to exploit object self-occlusions in pose estimation.
While all previous works directly regress the 3D object coordinates, DPOD~\cite{Zakharov_2019_ICCV} regresses a dense UV texture map, the UV values are further mapped to predefined 3D coordinates. EPOS~\cite{Hodan_2020_CVPR} divides object surface into fragments and predicts both fragments and local coordinates relative to the fragments. ZebraPose~\cite{Su_2022_CVPR} proposes a hierarchical binary vertex encoding defined by grouping surface vertices.
A segmentation network is employed to predict such codes in a coarse-to-fine manner.
2D-3D correspondences are thus extracted by predicting pixel-wise codes inside the object mask and mapping the codes to 3D coordinates.

\textbf{End-to-End Learning with PnP Solvers.}\quad
The PnP solver is typically treated as a non-differentiable layer, making it impossible to impose loss functions directly on the pose. Instead, surrogate loss functions are used in the first stage, encouraging the network to generate correct 2D-3D correspondences, but discarding information about the global object structure and the subsequent PnP step. Nevertheless, several attempts at differentiating through the PnP solver have been made. To this end, Brachmann \etal \cite{Brachmann_2017_CVPR} turn to numerical central differences, which introduce a computational burden. In their subsequent work~\cite{Brachmann_2018_CVPR}, the authors instead rely on approximations from the last step of the Gauss-Newton iterations.
MLP-based solvers~\cite{Di_2021_ICCV,Li_2019_ICCV,Hu_2020_CVPR,Wang_2021_CVPR}, where target pose is regressed from input geometry features are also deployed. Although these methods provide more flexibility on input modalities, it is hard to further improve their accuracy because of the lack of precise geometry model.
Chen \etal \cite{Chen_2020_CVPR} observe that the gradient of the optimal pose can be calculated by applying the implicit function theorem~\cite{krantz2002implicit} around the optimal solution. By contrast, EPro-PnP~\cite{Chen_2022_CVPR} relies on a loss function based on the whole posterior pose distribution instead of only its maximum, i.e., the true pose. However, as discussed in Section~\ref{sec:intro}, the averaging nature of PnP solvers makes loss functions based on the PnP solution suboptimal, as they will lead to gradients that degrade the accuracy of some of the 2D-3D correspondences. This is what we address in this paper.

\section{Method}
\subsection{Overview}
Let us now describe our method. To this end, we start with a general geometry-driven approach to 6D object pose estimation.
Given a single RGB image, the first stage of geometry-driven 6D pose estimation pipelines aims to predict $N$ noisy 2D-3D correspondences, potentially with associated weights. These can be expressed as
\begin{gather}
\bs{x}=\left[\bs{x}^T_1\quad\bs{x}^T_2\quad\cdots\quad\bs{x}^T_N \right]^T\in\mathbb{R}^{2N}\;, \\
\bs{z}=\left[\bs{z}^T_1\quad\bs{z}^T_2\quad \cdots \quad\bs{z}^T_N \right]^T\in\mathbb{R}^{3N}\;, \\
\bs{w}=\left[\bs{w}^T_1\quad\bs{w}^T_2\quad \cdots \quad\bs{w}^T_N \right]^T\in\mathbb{R}^{2N}\;,
\end{gather}
where $\{\bs{x}_i,\bs{z}_i,\bs{w}_i\}$ encodes the $i$-th correspondence with weight. 
Specifically, $\bs{z}_i$ is the 3D object point, $\bs{x}_i$ is its 2D projection, and $\bs{w}_i$ is the associated weight.
A PnP solver, compactly denoted by $g(\bs{x},\bs{z},\bs{w})$, can then be thought of as producing a maximum likelihood estimate of the pose $\bs{y}$, relying on the sum of the squared reprojection errors as a negative log likelihood (NLL). That is, we can write
\begin{equation}\label{eq:pnp}
    g(\bs{x},\bs{z},\bs{w}) = \mathop{\arg\min}_{\bs{y}}\frac{1}{2} \sum_i^N \left\Vert \bs{w}_i \circ \bs{r}_i \right\Vert^2\;,
\end{equation}
where
\begin{equation}
\bs{r}_i = \bs{x}_i - \pi(\bs{z}_i,\bs{y})
\end{equation}
is the reprojection residual for the $i$-th correspondence given pose $\bs{y}$, $\pi(\bs{z}_i,\bs{y})$ is the perspective projection involving the camera intrinsics, and $\circ$ denotes the element-wise product. 

If more than 4 correspondences are supplied, the PnP problem is over-determined, and the optimal solution can be thought of as a form of ``weighted average'' of the candidate poses obtained from all possible minimal correspondence sets.
When averaging is involved, 
penalizing the difference between the obtained pose and the ground truth does not guarantee to yield gradient directions that improve all the correspondences. In fact, 
Chen \etal~\cite{Chen_2020_CVPR} illustrated cases where the final pose has successfully converged to the ground truth while the correspondences had not.

To overcome this, we propose to introduce the ground truth \emph{before} solving the PnP problem and estimate the solution distribution around the ground-truth pose. This lets us build a loss function on top of this distribution, specifically, based on the distribution covariance. We discuss this in detail below.

\subsection{Covariance of the Pose Distribution}
\textbf{Linear Approximation of the PnP Solver.}\quad Let $\{\bs{x},\bs{z},\bs{w}\}$ denotes a set of noisy correspondences with weights, and $\bs{y}_{gt}$ be the ground-truth pose. Then, following a first-order Taylor expansion, the solution $\bs{y}$ obtained by a PnP solver $g(\bs{x},\bs{z},\bs{w})$ can be approximated as
\begin{equation}\label{eq:linearized}
\bs{y}=\bs{y}_{gt}+A(\bs{z},\bs{w})\cdot \bs{r}_{gt}\;, 
\end{equation}
where $\bs{r}_{gt}\in\mathbb{R}^{2N\times 1}$
is the residual vector given by $
\bs{r}_{gt}=\bs{x} - \bs{x}_{p}$, with $\bs{x}_{p,i}=\pi(\bs{z}_i,\bs{y}_{gt})$.
$A(\bs{z},\bs{w})$ encodes the gradient of pose $\bs{y}$ w.r.t. the perfect 2D locations $\bs{x}_p$, i.e.,
\begin{equation}
    A(\bs{z},\bs{w})=\dfrac{\partial \bs{y}}{\partial\bs{x}}
    =\left.\dfrac{\partial g(\bs{x},\bs{z},\bs{w})}{\partial\bs{x}}\right|_{\bs{x}=\bs{x}_p}.
\end{equation} 
which can be computed following the implicit function theorem~\cite{Chen_2020_CVPR, krantz2002implicit}.

\textbf{Pose Distribution and Covariance.}\quad 
Looking at Eq.~\ref{eq:linearized} reveals that, in our linearized model, the pose estimate is computed from a linear combination of the residuals in $\bs{r}_{gt}$. Let $M\in \mathbb{R}^{2N\times 2N}$ be the covariance matrix of the residuals. Then, the pose covariance matrix $C$ is given by
\begin{equation}\label{eq:cov}
    C=A(\bs{z},\bs{w})\cdot M\cdot A(\bs{z},\bs{w})^T\;,
\end{equation}
and its expected value is $\bs{y}_{gt}$.

This formalism still requires us to define the residuals covariance $M$. To this end,
we assume independence of the measurements, and express $M$ as the diagonal matrix
\begin{equation}\label{eq:rcov}
    M=diag\{\bs{r}_{gt}\circ\bs{r}_{gt}\}.
\end{equation}

During the calculation of the coefficient matrix $A(\bs{z},\bs{w})$, both the object 3D structure and the properties of the PnP solver are exploited. Furthermore, $A(\bs{z},\bs{w})$ is evaluated with perfect correspondences at the ground-truth pose $\bs{y}_{gt}$, eliminating the need to solve the PnP problem, since $\bs{y}_{gt}$ is the solution. This makes the covariance very fast to evaluate. It is worth noting that 3D coordinates vector $\bs{z}$ is treated as a constant during the calculation of the coefficient matrix $A(\bs{z},\bs{w})$, even if $\bs{z}$ is the output of the DNN's first stage, as is the case with dense prediction frameworks. This means that no gradient will be propagated from $A$ to $\bs{z}$.

\subsection{Linear-Covariance Loss}

\textbf{Pose Representation.}\quad 
Different representations of the 6D pose have been proposed in the past~\cite{Do2018LieNetRM,Li_2018_ECCV,Manhardt_2018_ECCV,Park_2020_CVPR,Zhou_2019_CVPR}.
The mathematical form of the pose covariance matrix in Eq.~\ref{eq:cov} is valid for any such representation. Nevertheless, to simplify the presentation of our loss function, we now discuss the representation that will be used in our experiments.

Specifically, we advocate for the use of a representation that reflects the evaluation metric of the target task. In particular, considering robotics applications where 3D error matters, we represent the pose with the transformed 3D coordinates of the object bounding box corners. That is, we write
\begin{equation}
\bs{y}=\left[T(\bs{b}_1;\bs{R},\bs{t})^T\quad\cdots\quad T(\bs{b}_8;\bs{R},\bs{t})^T\right]^T\in\mathbb{R}^{24}\;,
\end{equation}
where $\bs{b}_i$ is a 3D bounding box corner, and $T(\cdot;\bs{R},\bs{t})$ is rigid transformation performed with a pose encoded with a rotation matrix $\bs{R}$ and a translation vector $\bs{t}$, as would be output by a PnP solver. 

\textbf{Covariance Loss.}\quad  
Given the covariance matrix $C\in \mathbb{R}^{24\times 24}$, a natural choice of loss function is $trace(C)$, i.e., the sum of the diagonal covariance elements. This is because each diagonal element of the covariance matrix encodes the square error of the corresponding pose parameter, i.e., the square of its difference w.r.t. the ground-truth value. Such a square error, however, is sensitive to outliers, and does not reflect the nature of our pose representation, i.e., the fact that it contains 8 3D point coordinates and not 24 independent values.

Taking this into account, we therefore define our covariance loss as
\begin{equation}\label{eq:cov_loss}
    E_{cov}(\bs{w},\bs{r}_{gt})=\frac{1}{8}\sum_{i=1}^8\sqrt{\sum_{j=3i-2}^{3i} C_{jj}}\;,
\end{equation}
which we express as a function of $\bs{w}$ and $\bs{r}_{gt}$ to indicate that the gradient is computed w.r.t. only $\bs{w}$ and $\bs{r}_{gt}$, both of which depend on the network parameters, while all other quantities involved are treated as constants during back-propagation. 

\textbf{Weight-related Losses.}\quad 
Minimizing only the loss function of Eq.~\ref{eq:cov_loss} does not sufficiently constrain the weights $\bs{w}$, as for example, scaling them does not affect the solution of Eq.~\ref{eq:pnp}. Recall that, in linearly weighted least square problems, the weights are typically taken to be inversely proportional to the residual errors. Therefore, we propose to treat the weights as priors on the reprojection errors, and define a loss based on the covariance $C_{prior}$ of the pose prior.

One way to compute $C_{prior}$ would consist of re-defining the residuals as the inverse of the weights and reusing Eqs.~\ref{eq:rcov} and~\ref{eq:cov}. Instead, we approximate $C_{prior}$ as the inverse Hessian $H^{-1}$ of the NLL in Eq.~\ref{eq:pnp} at $\bs{y}_{gt}$ \cite{pawitan2001all}, i.e.,
\begin{equation}
C_{prior}=H(\bs{y}_{gt})^{-1}.
\end{equation}
This formulation is more accurate than using the inverse weights and more efficient since $H(\bs{y}_{gt})^{-1}$ is already evaluated when linearizing the PnP solver. 
We then define a prior loss as
\begin{equation} \label{eq:prior_loss}
    E_{prior}(\bs{w})=\frac{1}{8}\sum_{i=1}^8\sqrt{\sum_{j=3i-2}^{3i} C_{prior,jj}}\;.
\end{equation}

We further seek to supervise the weights to encourage them to benefit the final pose estimate obtained by the PnP solver. To achieve this, we rely on our linearized PnP solver and compute an error vector
\begin{equation}
    \bs{e}_{linear}(\bs{w})=\bs{y}-\bs{y}_{gt}=A(\bs{z},\bs{w})\cdot \bs{r}_{gt}\in \mathbb{R}^K \;.
\end{equation}
We then write a corresponding linear loss as
\begin{equation} \label{eq:linear_loss}
    E_{linear}(\bs{w})=\frac{1}{8}\sum_{i=1}^8\sqrt{\sum_{j=3i-2}^{3i} e^2_{linear,j}}\;.
\end{equation}

Note that the losses of Eqs.~\ref{eq:prior_loss} and~\ref{eq:linear_loss} are considered to be functions of $\bs{w}$ only, i.e., with the residuals $\bs{r}_{gt}$ detached from gradient computation.

\textbf{Linear-Covariance Loss.}\quad 
We write the final Linear-Covariance (LC) loss following a Laplace NLL formalism.
That is, we exploit the Laplace NLL loss, originating from the Laplace distribution and expressed as
\begin{equation}
    L_{nll}=\log(b)+\frac{\left|x-u\right|}{b},
\end{equation}
where $\left|x-u\right|$ is the error of $x$, and $b$ is a scale parameter acting as a prior prediction of the error. The Laplace NLL loss encourages a small error and a scale parameter equal to the error. In our context, it translates to the linear-covariance loss 
\begin{equation}
    L_{LC}=\log(E_{prior})+0.5\cdot\frac{E_{cov}+E_{linear}}{E_{prior}}\;.
\end{equation}
Ultimately, this loss seeks to minimize the sum of the covariance error and linear error, while encouraging the prior error to reflect this sum.

\section{Experiments}
We employ our linear-covariance loss in both dense and sparse correspondence-based methods. To this end, we use GDR-Net~\cite{Wang_2021_CVPR} as a first dense correspondence baseline, from which we build a competitive sparse correspondence baseline by replacing the GDR-Net output with 2D keypoint heatmaps. Furthermore, we exploit our linear-covariance loss in the ZebraPose~\cite{Su_2022_CVPR} framework, allowing us to achieve state-of-the-art performance.

\begin{figure}[t]
\begin{center}
\includegraphics[width=1\linewidth,trim=8 43 8 17]{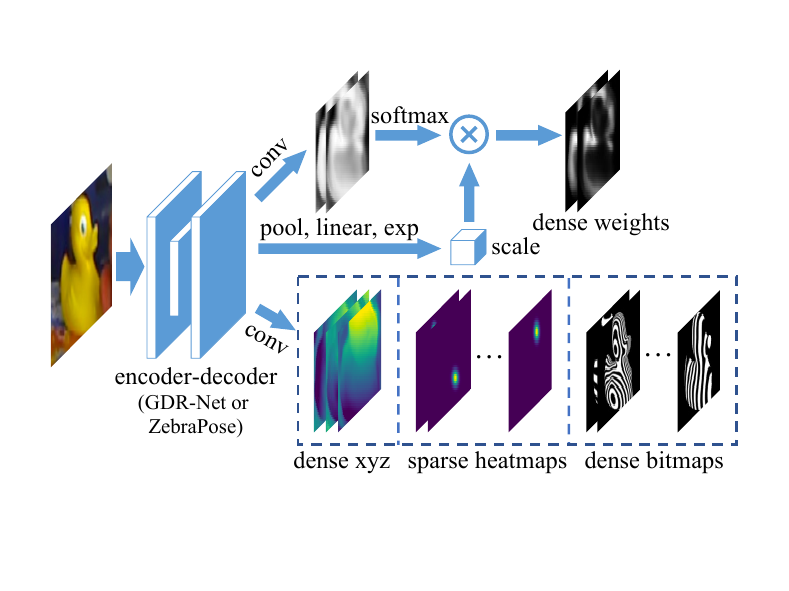}
\end{center}
   \caption{\textbf{Overall network structure.} Our experiments share a common pipeline, with differences in the detailed network structures and parameters. 
   We also regress dense visibility masks in the dense correspondence cases, which is omitted here for simplicity.}
\label{fig:structure}
\end{figure}
\subsection{Network Structure}
\label{subsec:scale-branch}
As illustrated by Fig.~\ref{fig:structure}, the networks used in our experiments, whether based on GDR-Net or ZebraPose, take as input a cropped image region within a detected target bounding box. An encoder-decoder network is used to extract geometry feature maps from this input patch. The specific network structures follow those of the baseline methods. We use the same structure as EPro-PnP~\cite{Chen_2022_CVPR} for dense weights regression, obtaining the weights using a scaled spatial Softmax. For methods whose loss functions cannot fully supervise the weights, we remove the scale branch, using the output of the Softmax as weights, which further constrains the weights to sum up to 1. For dense correspondence methods, the dense xyz or dense bitmaps are used for correspondence extraction with weights from the dense weights channel. The visibility mask channel is employed for correspondence selection.
For sparse correspondence methods, we only keep sparse heatmaps, each of which corresponds to a specific 2D keypoint. 
As the network acts on detected target image patches, 
we use the same setting of GDR-Net and Zebrapose, which is mainly based on Faster R-CNN~\cite{Girshick_2015_ICCV} and FCOS~\cite{Tian_2019_ICCV}.

\begin{figure}[t]
\begin{center}
\includegraphics[width=1\linewidth,trim=4 36 3.5 0]{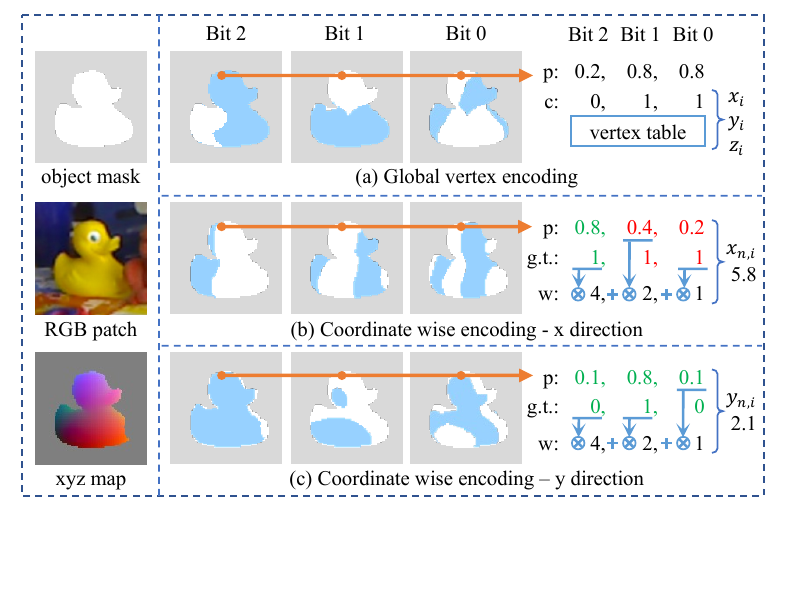}
\end{center}
   \caption{\textbf{Binary vertex encoding schemes.} (a) Global vertex encoding. Vertex codes are used as indices of the vertex table for coordinates lookup. (b) Coordinate wise encoding. Treating the vertex codes as normalized object coordinates, we keep the probability of the most significant mispredicted bit un-rounded and correct other bits before weighted summation. (c) If all predictions are correct, the least significant bit is kept un-rounded. In this figure, ``p'' is short for ``probability'', ``c'' is ``code'', ``w'' is ``weight'', and ``g.t.'' is ``ground-truth''. We omit the z direction for simplicity.}
\label{fig:binary_encoding}

\end{figure}

\subsection{Implementation Details}
\textbf{Correspondence Extraction.}\quad
With GDR-Net, dense correspondences can be directly obtained from the 3D coordinates and associated weights output by the network. When weights are not available, the correspondences are selected based on the visibility mask.
In the sparse case, keypoint locations and their standard deviations are estimated from the heatmaps with a DSNT~\cite{nibali2018numerical} layer. We use the inverse of the standard deviations of the 2D keypoint predictions as weights.

\textbf{Coordinate-wise Binary Encoding.}\quad As illustrated in Fig.~\ref{fig:binary_encoding}~(a), the original ZebraPose recursively subdivides the object surface $N$ times in a coarse-to-fine manner, generating $N$ binary bitmaps. For any target pixel, a binary code of length $N$ is obtained by concatenating the binary prediction in each bitmap. The vertex 3D coordinates are then retrieved by indexing a predefined vertex table with the binary codes.
 
As the lookup operation is not differentiable. We instead assign a code for each component of the vertex 3D coordinates, the code is directly treated as normalized coordinate of the component. Specifically, a coordinate $c \in \{x,y,z\}$ in the range $[c_{min}, c_{max}]$ is normalized to $c_n$ in $[0,2^{M_c}-1]$, i.e., to the representable range of $M_c$ bit integers, via a linear transform.
The binary representation of integer $round(c_n)$ is directly used as the binary encoding of $c$. As a result, a single vertex is expressed with 3 binary codes, with no lookup operation involved.

To recover the normalized coordinates with available ground-truth, as illustrated by Fig.~\ref{fig:binary_encoding}~(b), instead of directly computing the weighted sum of predicted bits based on their significance, we first check their correctness and find the most significant mispredicted bit.
This bit is substituted with the predicted unrounded probability. All other mispredicted bits are corrected before weighted summation, since their errors are negligible compared to the most significant erroneous bit. If all bits are correct or no ground truth available, the probability of the least significant bit is kept unrounded. In this way, the coordinates are differentiable, at the cost of more bits to predict, and the benefits of binary prediction are preserved. 

\textbf{Training Details.}\quad Following the baselines' strategy, we train a separate model from an ImageNet-pretrained backbone for each object with same optimizer, scheduler, and training epoch or step count as the baseline methods for fair comparison. The original loss functions for correspondence learning are kept.
Specifically, GDR-Net~\cite{Wang_2021_CVPR} penalizes the L1 difference between a regressed normalized 3D coordinate map $\hat{M}_{XYZ}$ and the ground truth $M_{XYZ}$ within the visible region $M_{vis}$. We use the binary cross entropy loss to train the visible region estimate $\hat{M}_{vis}$ and use $\hat{M}_{vis}$ for correspondence extraction if no weights are available. The full loss for the GDR-Net baseline is
\begin{equation}
\begin{split}
    L_{GDR-base}=&\Vert(\hat{M}_{XYZ}-M_{XYZ})\odot M_{vis}\Vert_1\\
     & + \alpha_{GDR}\cdot BCE(\hat{M}_{vis}, M_{vis})\;,
\end{split}
\end{equation}
with $\alpha_{GDR}=0.25$ and $\odot$ denoting the element-wise product.
For ZebraPose~\cite{Su_2022_CVPR}, the original losses for mask prediction and binary code learning are both kept, giving
\begin{equation}
    L_{Zebra-base}=L_{mask}+\alpha_{Zebra} \cdot L_{hier}\;,
\end{equation}
in which $L_{mask}$ is the L1 loss for object mask prediction, $L_{hier}$ is a hierarchical loss for binary code learning and $\alpha_{Zebra}=3$.
The full loss when our LC loss is applied then is
\begin{equation}
    L_{i-full}=L_{i-base}+\beta_{i} \cdot L_{LC}, \;i\in\{GDR, Zebra\}.
\end{equation}
For the experiments in Sec.~\ref{subsec:cmp_sota}, we use $\beta_{GDR}=0.02$ and $\beta_{Zebra}=0.03$.

The pose loss is fully applied shortly after training begins since the network generates random correspondences at the very start. As matrix inversion is involved during linearization, the loss may generate large gradients in corner cases. We implement a gradient clipper which tracks the  magnitude of the gradients and clip overly large ones. For outlier robustness, we employ the Huber function~\cite{huber} adaptively when evaluating the squared residuals in Eq.~\ref{eq:rcov} and the squared weights involved in the linearization process. 

\textbf{Efficient Loss Computation.}\quad For efficiency, we compute the covariance matrix in the most compact 6D representation and transform it to our target representation. For a specific pose representation $\bs{y}^K$ with $K$ components, which is transformed from its 6D version $\bs{y}^6$ as $\bs{y}^K=f(\bs{y}^6)$, we calculate the Jacobian matrix $J$ of $f$ w.r.t. $\bs{y}^6$. Given the covariance matrix $C^{6\times 6}$ of $\bs{y}^6$, the covariance of $\bs{y}^K$ can be calculated as $C^{K\times K} = J\cdot C^{6\times 6}\cdot J^T$; only  its diagonal elements are evaluated to compute the loss.
Given $N$ correspondences, the complexity of our loss can be reduced from $O(K^2 N)+O(K^3)$ to $O(6^2 N)+O(6^3)+O(K)$.
On a single NVIDIA A100 GPU, with the LC loss, the GDR-Net based network takes around 3 hours on LM-O and 1 hour on YCB-V for a single object, and the ZebraPose based network takes around 24 hours for a single object on both datasets.

\subsection{Datasets and Metrics}
\textbf{Datasets.}\quad We evaluate our approach on the widely used Linemod-Occluded (LM-O)~\cite{LMO-2014} and YCB-Video (YCB-V)~\cite{xiang2018posecnn} datasets. LM-O is an extension of the Linemod (LM)~\cite{linemod} dataset. It has a total of 1214 images annotated for 8 objects under severe occlusion and is only used for testing. About 1.2k images for each object from the LM dataset are used as real training images. YCB-V is a large challenging video dataset containing about 133k images with strong occlusions and clutter. It contains 21 objects, some of which are symmetric or texture-less. We also adopt the publicly available physically-based rendering (pbr)~\cite{Denninger2023} images for training.

\textbf{Metrics.}\quad We employ the widely used ADD(-S)~\cite{linemod} score to compare with other methods. Under this metric, a pose is correct if the average distance between the object vertices transformed by the predicted pose and by the ground-truth pose is lower than a threshold. \mbox{ADD-S} is used for symmetry objects, replacing the average distance by the average nearest distance. The ADD(-S) score is computed with the threshold corresponding to 10\% of the object diameter. For the YCB-V dataset, we also report the AUC (area under the curve) of the ADD(-S), varying the threshold up to a maximum of 10 cm. 

As the ADD(-S) metric is defined in 3D space, in our ablations, we also report the Average Recall of Maximum Symmetry-Aware Projection Distance (MSPD) $AR_{MSPD}$, a metric from the BOP benchmark~\cite{Hodan_2018_ECCV} based on 2D reprojection errors.

\subsection{Comparison with the State of the Art}
\label{subsec:cmp_sota}
\begin{table}
\begin{center}
\begin{tabular}{c|c|c}
\hline
Method & Training Data & ADD(-S) \\
\hline
RePOSE~\cite{Iwase_2021_ICCV}& real+syn & 51.6\\
RNNPose~\cite{Xu_2022_CVPR}& real+syn & 60.65\\
SO-Pose~\cite{Di_2021_ICCV}& real+pbr & 62.3\\
DProST~\cite{dprost_2022_eccv} & real+pbr & 62.6 \\
\hline
GDR-Net~\cite{Wang_2021_CVPR}& real+pbr & 62.2\\
ZebraPose~\cite{Su_2022_CVPR}& real+pbr& 76.9 \\
\hline
GDR-Net-LC & real+pbr& 66.48 \\
ZebraPose-LC & real+pbr& \textbf{78.06} \\
\hline
\end{tabular}
\end{center}

\caption{\textbf{Comparison with the state of the art on LM-O.} The ``LC'' postfix indicates the LC loss is applied.}
\label{tab:lmo-sota}

\end{table} 
\begin{table}
\begin{center}
\begin{tabular}{c|c|c|c}
\hline
Method & ADD(-S) &\makecell{AUC of\\ADD-S}&\makecell{AUC of\\ADD(-S)}\\
\hline
RePose~\cite{Iwase_2021_ICCV} & 62.1 & 88.5 & 82.0\\
RNNPose~\cite{Xu_2022_CVPR} & 66.4 & - & 83.1\\
SO-Pose~\cite{Di_2021_ICCV} & 56.8 & *90.9 & *83.9 \\
DProST~\cite{dprost_2022_eccv} & 65.1 & - & 77.4\\
\hline
GDR-Net~\cite{Wang_2021_CVPR}  & 60.1 & *91.6 & *84.4 \\
ZebraPose~\cite{Su_2022_CVPR}& 80.5 & 90.1 & 85.3\\
\hline
GDR-Net-LC & 70.6 & 89.8, *94.1 & 84.0, *88.8 \\
ZebraPose-LC & \textbf{82.4} & \textbf{90.8}, \textbf{*95.0} &\textbf{86.1}, \textbf{*90.8}\\
\hline
\end{tabular}
\end{center}

\caption{\textbf{Comparison with the state of the art on YCB-V.} * indicates that the AUC is calculated with 11-points interpolation.}
\label{fig:ycbv-sota}
\end{table} 

In this section, we compare the performance obtained when applying our loss to GDR-Net~\cite{Wang_2021_CVPR} and ZerbraPose~\cite{Su_2022_CVPR} with other start-of-the-art methods. We also compare our loss function with the state-of-the-art differentiable PnP methods, namely, BPnP~\cite{Chen_2020_CVPR} and EPro-PnP~\cite{Chen_2022_CVPR}. 

\textbf{Results on LM-O.}\quad We report the ADD(-S) score on LM-O in Tab.~\ref{tab:lmo-sota}.
Applying the LC loss on GDR-Net surpasses most methods, and we achieve state-of-the-art performance when applying it to ZebraPose.

\begin{figure}[t]
\begin{center}
\includegraphics[width=1\linewidth,trim=0 50 0 0]{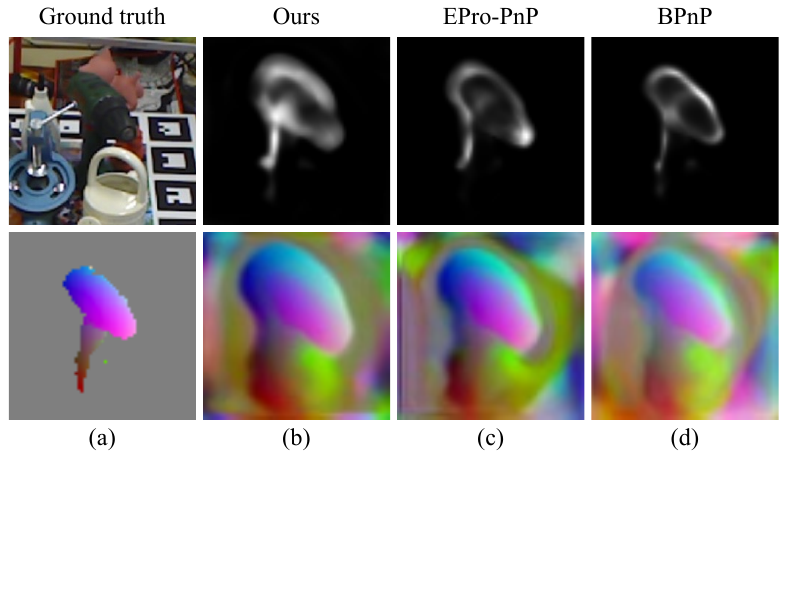}
\end{center}
   \caption{\textbf{Visualization of different PnP layers.} In the first column, we show the input image patch in the first row and the ground-truth object coordinates in the second row. The remaining images in the first row visualize the normalized weight maps for different methods. The remaining images in the second row visualize the predicted object coordinates corresponding to the weights in the first row.}
\label{fig:pnpcmp}
\end{figure}
 
\begin{table}
\begin{center}
\begin{tabular}{c|c|c|c}
\hline
Method & ADD(-S) &Correctness & Runtime \\
\hline
BPnP~\cite{Chen_2020_CVPR} & 64.1 & 53.9 & $\sim$ 30 ms\\
EPro-PnP~\cite{Chen_2022_CVPR} & 64.5 & 59.3 & $\sim$ 80 ms \\
Ours & \textbf{66.5} & \textbf{99.9} &$\sim$ \textbf{15} ms\\
\hline
\end{tabular}
\end{center}

\caption{\textbf{Comparison between PnP layers on LM-O.}}
\label{tab:pnp}
\end{table}
 
\textbf{Results on YCB-V.}\quad
As summarized in Tab.~\ref{fig:ycbv-sota}, applying the LC loss on GDR-Net produces results second only to ZebraPose. Furthermore, we achieve state-of-the-art performance when the LC loss is applied to ZebraPose. We implement a symmetry-aware training scheme which selects the ground-truth pose on the fly based on the average distance between the predicted 3D coordinates at randomly selected 2D locations and possible 3D coordinates at the same locations under the symmetric ground-truth pose. This scheme is only applied to the GDR-Net-based experiments on YCB-V, including the baseline and the model with our LC loss for fair comparison.

\textbf{Comparison with Differentiable PnP layers.}\quad As summarized in Tab.~\ref{tab:pnp}, we carry out experiments on LM-O with the GDR-Net baseline, and compare the methods based on several metrics 
including the ADD(-S) score, the gradient correctness and the runtime per training step, 
the correctness and runtime are evaluated at the end of training. Note that BPnP~\cite{Chen_2020_CVPR} does not fully constrain the weights, thus we remove the scale branch as stated in Sec.~\ref{subsec:scale-branch}. Our method yields the best \mbox{ADD(-S)} score on \mbox{LM-O}. More importantly, it generates a much larger percentage of correct gradients. A 3D point is considered to have correct gradients if moving in the negative gradient direction leads to a smaller 2D reprojection error. A pose loss yielding a higher gradient correctness provides more consistent supervision signals. 
The consistency is reflected by the dilated weight and coordinate maps shown in Fig.~\ref{fig:pnpcmp}, in particular by looking at pixels outside of but close to the actual object region.
Such pixels receive supervision only from the pose loss and thus indicate the differences between the different pose losses. 
Higher correctness helps the network to predict correct correspondences for such pixels. 
This virtually expands the target object size in 3D object space and in 2D image space, which facilitates better pose estimates. The LC loss yields 99.9\%  gradient correctness, generating the most dilated maps. By contrast, the other losses have weaker consistency and thus tend to predict less accurate correspondences in these regions. Finally, as shown in Tab.~\ref{tab:pnp}, our LC loss yields the fastest runtime, evaluated on an NVIDIA A100 GPU with a batch size of 32. This is due to our linearization of the PnP solver, removing the need for an iterative solution.

\subsection{Ablation Study on LM-O}

We evaluate the influence of each component of our LC loss by applying it to the dense correspondence-based GDR-Net and to the sparse correspondence-based one on the LM-O dataset. 
The results are summarized in Tab.~\ref{tab:ab-dense} and Tab.~\ref{tab:ab-sparse}.

\textbf{Effectiveness of the Pose Representation.}\quad We validate the ability of the LC loss at integrating different application preferences by switching to a pose representation defined in the 2D image space. Similarly to the 3D case, a pose representation $\bs{y}^{2D}\in \mathbb{R}^{16}$ can be defined as the projected 2D coordinates of the 8 object bounding box corners.
Switching from the LC loss in 3D space to its 2D space counterpart (B0~\textit{vs.}~B1) causes a slight ADD(-S) drop. We further compare the MSPD scores between these two versions.
The network trained with 2D space LC loss yields an $AR_{MSPD}$ score of 84.14, better than the 83.98 of the 3D case. This validates the effectiveness of using a metric aware pose representation for different applications.

\begin{figure}[t]
\begin{center}
\includegraphics[width=1\linewidth,trim=0 50 0 0]{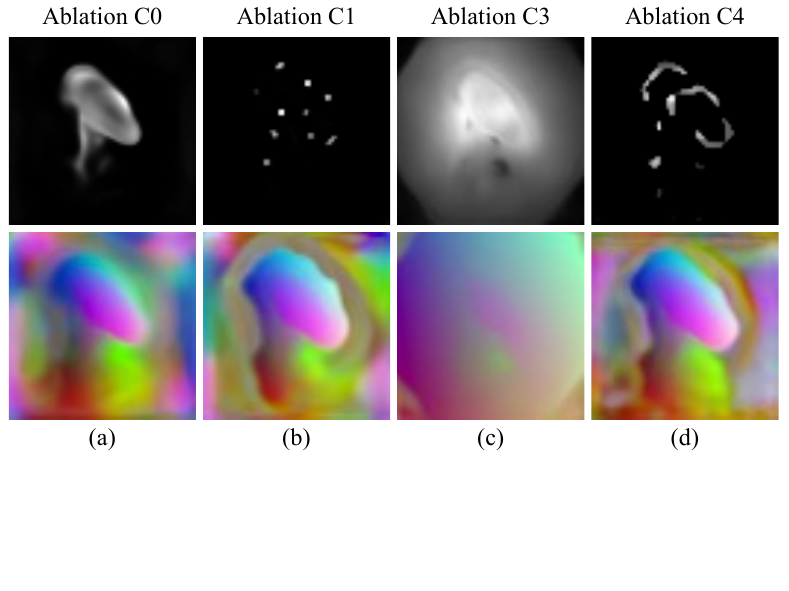}
\end{center}
   \caption{\textbf{Visualizations for our different ablations.} We show the visualizations of the same input patch as in Fig.~\ref{fig:pnpcmp}. The first row visualizes the normalized weight maps for different settings. The second row shows the predicted object coordinates corresponding to the weights.}
\label{fig:ablation}
\end{figure}

\textbf{Effectiveness of the Covariance Loss.}\quad
The covariance loss is designed for both correspondence learning and weight learning. It aims to minimize the residuals of the correspondences, and encourages the weights to be inversely proportional to the residuals. 
We investigate its effectiveness by detaching, in turn, the residuals $\bs{r}_{gt}$ and weights $\bs{w}$ from the covariance loss. When the residuals are detached (B0~\textit{vs.}~C0), the only loss for correspondence learning is the original surrogate loss, which lacks supervision on the background pixels. Thus the learned weights and coordinates are restricted to the visible part of the target. When the weights are detached (B0~\textit{vs.}~C1), the network yields over-concentrated weights lying outside of the object region.
This is evidenced by Fig.~\ref{fig:ablation}~(b); although the network learns these emphasized correspondences correctly, its performance still degrades compared to our approach. If the covariance loss is entirely removed (B0~\textit{vs.}~C2), the pipeline suffers from a large performance drop.

\begin{table}
\begin{center}
\begin{tabular}{c|l|c}
\hline
Row & Method & ADD(-S)\\
\hline
A0 & GDR-Net baseline & 59.29 \\
A1 & GDR-Net~\cite{Wang_2021_CVPR}  & 62.2 \\
\hline
B0 & A0 + 3D LC loss & \textbf{66.48}  \\
B1 & A0 + 2D LC loss & 65.99 \\
\hline 
C0 & B0 + detach residual from $E_{cov}$ & 61.03 \\
C1 & B0 + detach weights from $E_{cov}$ & 65.23 \\
C2 & B0 + remove $E_{cov}$ & 45.61 \\
C3 & B0 + remove $E_{linear}$ & 65.82 \\
C4 & B0 + remove $E_{prior}$ & 60.69 \\
\hline
\end{tabular}
\end{center}

\caption{\textbf{Ablation study of the dense correspondence-based method on LM-O.} Block~A: Comparison of the baseline method and the original method. Block~B: Comparison between losses derived from different pose representations. Block~C: Ablations of each component of the linear-covariance loss on the dense correspondence-based baseline.}
\label{tab:ab-dense}

\end{table} 
\textbf{Effectiveness of the Linear Loss.}\quad The linear loss is a linear approximation of the actual pose error. As it acts on the weights, it seeks to emphasizes the correspondences that are more beneficial for the pose. As shown in Tab.~\ref{tab:ab-dense}, after removing $E_{linear}$ from the LC loss, the ADD(-S) score drops slightly (B0~\textit{vs.}~C3). However, this makes a significant difference on the learned coordinate and weight maps; as illustrated in Fig.~\ref{fig:ablation}~(c), when trained without linear loss, the network tends to confidently extrapolate the 3D coordinates to pixels far away from the target region or occluded. Such extrapolated coordinates are unreliable compared to correspondences near or inside object region, and thus they are suppressed by the linear loss.

\textbf{Effectiveness of the Prior Loss.}\quad
Together with the other loss terms, the prior loss fully constrains the weights, and allows for fine-grained weights learning.  When the prior loss is removed (B0~\textit{vs.}~C4), we also remove the scale branch as stated in Sec.~\ref{subsec:scale-branch}. As illustrated by \mbox{Fig.~\ref{fig:ablation}~(d)}, the network also learns concentrated weights but with a larger performance drop.

\begin{table}
\begin{center}
\begin{tabular}{c|c|c|c}
\hline
Row & Training Loss & Solver & \makecell{ADD(-S)\\}\\
\hline
0 & MLE & PnP RANSAC & 57.89 \\
1 & MLE & PnP weighted & 59.90 \\
2 & MLE + LC loss & PnP RANSAC & 57.74 \\
3 & MLE + LC loss & PnP weighted & \textbf{61.08} \\
\hline
\end{tabular}
\end{center}

\caption{\textbf{Ablation for the sparse correspondence-based method on the LM-O dataset.} }
\label{tab:ab-sparse}
\end{table} 
\textbf{Effectiveness with a Sparse Correspondence-based Method.}\quad In the dense case, the weights serve as both attention mechanism, emphasizing some important or stable points during training, and indicators for well learned correspondences during testing. The sparsity in sparse correspondence methods limits the attention feature. The loss function in sparse cases mostly encourages the network to predict better weights. Our sparse baseline is trained with a Laplace MLE loss, similar to the Gaussian MLE loss in~\cite{Merrill_2022_CVPR}. The predicted standard deviations are encouraged to capture point location errors, and their inverse are subsequently used as weights in the PnP solver. As shown in Tab.~\ref{tab:ab-sparse}, applying our loss in this scenario also brings a performance gain. As the networks yield very close performance when not using weights (Row~0~\textit{vs.}~Row~2), this gain comes mostly from better weights learning.
Note that the \emph{PnP RANSAC} solver does not use weights but uses RANSAC~\cite{fischler1981random} to evict outliers, thus reflecting only the quality of 2D point locations. The \emph{PnP weighted} solver iteratively solves Eq.~\ref{eq:pnp} using the \emph{PnP RANSAC}'s solution as starting point, which effectively relies on predicted weights to evict outliers, reflecting the quality of the predicted weights.

\textbf{Experiments on other pose representations and other NLL formulations.}\quad We also carried out experiments on losses based on quaternions, axis-angles and two-column~\cite{Zhou_2019_CVPR} pose representations on LM-O.
For the Laplace NLL formulation, we used the sum of the square roots of the covariance diagonal for $\{E_{cov}, E_{prior}\}$ and the sum of the absolute values of the approximate error for $E_{linear}$. 
To further validate the distribution from which the NLL loss originates, we switched from Laplace to Gaussian, using the trace of the covariance and the sum of the squared errors to build the losses.
As shown in Tab.~\ref{tab:exp-tab}, the performance is robust to the pose representation. However, the Laplace NLL losses yield much better results than their Gaussian counterparts.
\begin{table}
\begin{center}
\setlength\tabcolsep{4pt}
\begin{tabular}{c|c|c|c|c}
\hline
dist.\verb|\| rep.  & ours-3D & quater. & axis-ang. & two-col.~\cite{Zhou_2019_CVPR}\\
\hline
Laplace & 66.48 & 66.72 & 65.85 & 66.39 \\
Gaussian & 61.90 & 60.72 & 59.79 & 60.37 \\
\hline

\end{tabular}
\end{center}

\caption{\textbf{Results on different pose representations and NLL distributions in ADD(-S) score.}}

\label{tab:exp-tab}
\end{table} 

\section{Conclusion}
In this work, we have proposed a linear-covariance loss for end-to-end weights and correspondence learning of two stage geometry-based 6D pose estimation networks. 
This new loss function addresses the problem originating from the averaging nature of PnP solvers, resulting in gradients that may seek to degrade the accuracy of some correspondences.
At the heart of our loss is the idea of introducing ground-truth information by linearizing the PnP solver around the ground truth before the pose is actually solved, and building the loss function for correspondence learning on the covariance of the pose distribution. Our extensive experiments have validated the effectiveness of our LC loss on both sparse and dense correspondence-based methods and on two standard benchmarks. 

Nevertheless, the LC loss cannot learn correspondences from scratch, and a surrogate loss function to supervise the correspondences remains necessary for providing the LC loss with an initial object 3D structure. In the future, we will seek to apply our loss function to category-level object pose estimation~\cite{Chen_2021_CVPR,Di_2022_CVPR,Wang_2019_CVPR}, where precise object structure is not available.

\ificcvfinal

\noindent \textbf{Acknowledgement.}\quad This work was supported in part by China Scholarship Council (CSC) Grant 202006020218, and in part by the National Natural Science Foundation of China (NSFC) under Grant 52127809 and Grant 51625501.
\fi

\section{Supplementary}
\subsection{Linearization of PnP Solver}
\noindent \textbf{Implicit Function Theorem.}
The implicit function theorem (IFT)~\cite{krantz2002implicit} states the following:

Given \mbox{$f:\mathbb{R}^{n+m}\to \mathbb{R}^m$} a continuously differentiable function
 with input $(\bs{a},\bs{b})\in \mathbb{R}^n\times\mathbb{R}^m$, if a point $(\bs{a}^*,\bs{b}^*)$ satisfies
\begin{equation}\label{eq:ift_f}
    f(\bs{a}^*,\bs{b}^*)=\bs{0}\;,
\end{equation}
and the Jacobian matrix $\frac{\partial f}{\partial\bs{b}}(\bs{a}^*,\bs{b}^*)$ is invertible, then there exists a unique continuously differentiable function \mbox{$g(\bs{a}):\mathbb{R}^n\to\mathbb{R}^m$} such that
\begin{equation}\label{eq:ift_g}
    \bs{b}^*=g(\bs{a}^*)\;,
\end{equation}
and
\begin{equation}\label{eq:ift_fg}
f(\bs{a}^*,g(\bs{a}^*))=\bs{0}\;.
\end{equation}
The Jacobian matrix $\frac{\partial g}{\partial\bs{a}}(\bs{a}^*)$ is given by
\begin{equation}\label{eq:ift_grad}
\frac{\partial g}{\partial\bs{a}}(\bs{a}^*)=-\left[ \frac{\partial f}{\partial \bs{b}}(\bs{a}^*,\bs{b}^*) \right]^{-1}\cdot
\frac{\partial f}{\partial \bs{a}}(\bs{a}^*,\bs{b}^*)\;.
\end{equation}

\noindent \textbf{PnP Linearization.}
Following the same notation as in the main paper, the PnP solver computes the function
\begin{equation}\label{eq:sup_pnp}
    g(\bs{x},\bs{z},\bs{w}) = \mathop{\arg\min}_{\bs{y}}\frac{1}{2} \sum_i^N \left\Vert \bs{w}_i \circ \bs{r}_i \right\Vert^2\;,
\end{equation}
where $\bs{x}_i$ is the $i$-th image 2D point, $\bs{z}_i$ is the $i$-th 3D point, $\bs{w}_i$ is the corresponding weight, and
\begin{equation}
\bs{r}_i = \bs{x}_i - \pi(\bs{z}_i,\bs{y})
\end{equation}
is the reprojection residual for the $i$-th correspondence given pose $\bs{y}$.

Eq.~\ref{eq:sup_pnp} implies that the solution $\bs{y}^*$ is the stationary point of the negative log likelihood (NLL) function
\begin{equation}\label{eq:nll}
    nll(\bs{y}) = \frac{1}{2} \sum_i^N \left\Vert \bs{w}_i \circ \bs{r}_i \right\Vert^2\;.
\end{equation}

Since $\bs{y}^*$ is the stationary point of the NLL function, the first order derivative of the NLL w.r.t. $\bs{y}^*$ should be zero, i.e.,
\begin{equation}
    \left.\frac{\partial nll(\bs{y})}{\partial\bs{y}}\right|_{\bs{y}=\bs{y}^*}=\bs{0}\;.
\end{equation}

Eqs.~\ref{eq:ift_f},~\ref{eq:ift_g} and ~\ref{eq:ift_fg} in the PnP case can subsequently be specialized as
\begin{equation}
f(\bs{x},\bs{y},\bs{z},\bs{w})|_{\bs{y}=\bs{y}^*}=\left.\frac{\partial nll(\bs{y})}{\partial\bs{y}}\right|_{\bs{y}=\bs{y}^*}=\bs{0}\;,
\end{equation}
\begin{equation}
\bs{y}^*=g(\bs{x},\bs{z},\bs{w})\;,
\end{equation}
and
\begin{equation}
f(\bs{x},g(\bs{x},\bs{z},\bs{w}),\bs{z},\bs{w})|_{\bs{y}=\bs{y}^*}=\bs{0}\;.
\end{equation}

According to Eq.~\ref{eq:ift_grad}, the gradient of the pose $\bs{y}$ w.r.t. the 2D locations $\bs{x}$ at $\bs{y}^*$ is
\begin{equation}\label{eq:dydx}
\begin{split}
    \left.\frac{\partial\bs{y}}{\partial\bs{x}}\right|_{\bs{y}^*}&=
    \left.\frac{\partial g(\bs{x},\bs{z},\bs{w})}{\partial \bs{x}}\right|_{\bs{y}^*}\;,\\
    &=\left.-\left[\left[\frac{\partial^2 nll(\bs{y})}{\partial \bs{y}^2}\right]^{-1} \cdot
    \frac{\partial^2 nll(\bs{y})}{\partial \bs{y}\partial\bs{x}}\right]\right|_{\bs{y}*}\;,\\
    &=\left.-H^{-1}\cdot
    \frac{\partial^2 nll(\bs{y})}{\partial \bs{y}\partial\bs{x}}\right|_{\bs{y}^*}\;,
\end{split}
\end{equation}
with $nll(\bs{y})$ defined by Eq.~\ref{eq:nll}.

Given the noisy correspondences $\{\bs{x},\bs{z},\bs{w}\}$, we compute the perfect correspondences $\{\bs{x}_p,\bs{z},\bs{w}\}$ with $\bs{x}_{p,i}=\pi(\bs{z}_i,\bs{y}_{gt})$ under the ground-truth pose $\bs{y}_{gt}$. We then linearize the PnP solver around $\{\bs{x}_p,\bs{z},\bs{w}\}$ and $\bs{y}_{gt}$ using the first-order Taylor expansion as
\begin{equation}
    \bs{y}=\bs{y}_{gt}+A(\bs{z},\bs{w})\cdot \bs{r}_{gt}\;,
\end{equation}
with
\begin{equation}
\bs{r}_{gt} = \bs{x}-\bs{x}_{gt}
\end{equation}
being the residual vector at $\bs{y}_{gt}$,
and 
\begin{equation}
    A(\bs{z},\bs{w})=\left.-H^{-1}\cdot
    \frac{\partial^2 nll(\bs{y})}{\partial \bs{y}\partial\bs{x}}\right|_{\bs{y}=\bs{y}_{gt},\bs{x}=\bs{x}_p}.
\end{equation}
The Hessian $H$ of the NLL function is also used to compute the prior loss, as stated in Sec.~3.3 in the main paper.

\begin{figure}[t]
\begin{center}
\includegraphics[width=0.9\linewidth,trim=0 20 0 1]{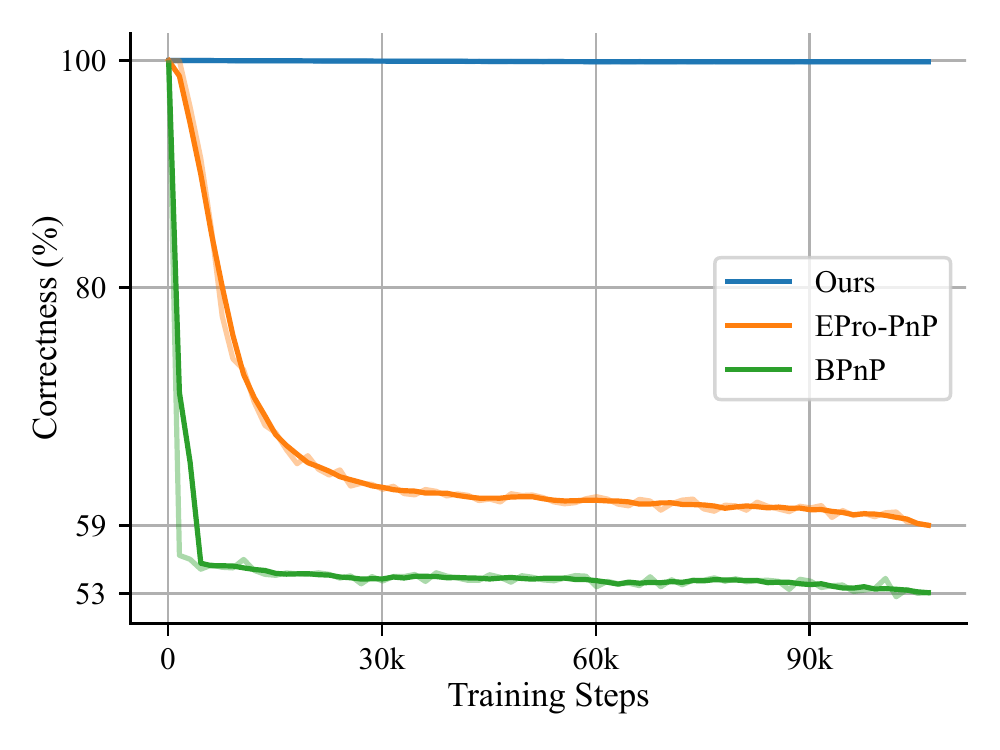}
\end{center}
   \caption{\textbf{Correctness curves of the PnP layers.}  A 3D point is considered to have a correct gradient if moving in the negative gradient direction leads to a smaller 2D reprojection error. The LC loss yields almost 100\% correctness. The correctness of EPro-PnP drops slowly, and ending with about 59\% correctness. BPnP drops quickly when training begins, and ends with about 53\% correctness. The dark curves are smoothed versions of the light ones.}
\label{fig:corr}
\end{figure}

\subsection{Detailed Results on Gradient Correctness}
We further provide the whole correctness curves to show how the correctness evolves as training progresses. 

As illustrated in Fig.~\ref{fig:corr}, at the very beginning, when the correspondences have large errors, both EPro-PnP~\cite{Chen_2022_CVPR} and BPnP~\cite{Chen_2020_CVPR} have good correctness. However, their correctness drops when the training proceeds. Since the linear-covariance loss is designed to address this problem, it always maintains a correctness close to 100\%.

\subsection{Details on ZebraPose-based Experiments}
\noindent \textbf{Implementation Details.}
\begin{table}[t]
\begin{center}
\begin{tabular}{c|l|c}
\hline
Row & Method & ADD(-S)\\
\hline
A0 & ZebraPose~\cite{Su_2022_CVPR} & 76.91 \\
A1 & ZebraPose baseline & 75.19 \\
A2 & A1 + LC loss & \textbf{78.06} \\
\hline
\end{tabular}
\end{center}
\caption{\textbf{Results of the ZebraPose~\cite{Su_2022_CVPR} based experiments on the LM-O dataset.}}
\label{tab:sup-zebra}
\end{table}
Our coordinate-wise encoding scheme assigns 3 binary codes to a vertex, eliminating the look up operation. To reduce the number of binary bits for prediction, we rotate some of the objects to minimize their span along the $x,y,z$ directions. We use 7 bits to represent the coordinate component with the largest span, and calculate the binary count of the other components based on their relative span w.r.t. largest one. 
Specifically, given the sizes $s_i, i\in\{x,y,z\}$, of an object and their maximum $s$, the bit count of each component is calculated as $n_i = \mathrm{round}( n + log_2(s_i/s))$, where $n = 7$ is the maximum bit count per component. This is to reduce the unpredictable bits for flat-shaped objects such as scissors.

\begin{figure}[t]
\begin{center}
\includegraphics[width=1\linewidth,trim=0 30 0 0]{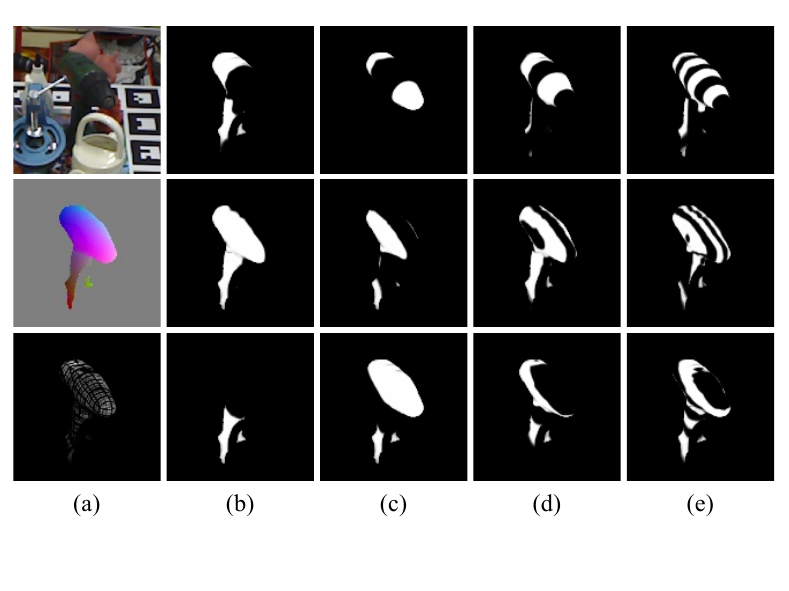}
\end{center}
   \caption{\textbf{Visualizations for the ZebraPose-based model.}  (a) Visualizations of the input image patch, decoded object coordinates and the predicted weight map. (b)-(e) Visualizations of the predicted masks of coordinate components with the most significant bit at the left and the $x$ component at the top. The pixels predicted as background are masked out for clarity.}
\label{fig:zebra}
\end{figure}
\noindent \textbf{Results.}
As shown in Tab.~\ref{tab:sup-zebra}, after switching from the global vertex encoding to our coordinate-wise encoding (A0~\textit{vs.}~A1), the performance drops by about 1.7 points. When the LC loss is applied, the performance drop is compensated, surpassing the original ZebraPose~\cite{Su_2022_CVPR}.

\noindent \textbf{Visualizations.}
As illustrated by Fig.~\ref{fig:zebra}, the learned weight map successfully captures the error distribution of the predicted 3D coordinates in a geometry-aware manner, generating low weights for code transition regions and high weights for object endpoint regions.

\subsection{Detailed Results on LM-O and YCB-V}
For the LM-O dataset, we provide the detailed comparison of ADD(-S) scores with state-of-the-art methods, when the linear-covariance (LC) loss is applied to GDR-Net and ZebraPose on LM-O in Tab.~\ref{tab:lmo-detail}.

For the YCB-V dataset, we provide the detailed comparison of ADD(-S) scores (Tab.~\ref{tab:ycbv-detail-add}) and AUC scores (Tab.~\ref{tab:ycbv-detail-auc}) between the baseline methods and the versions where the LC loss is applied.

\begin{table}
\begin{center}
\scalebox{0.92}{
\begin{tabular}{l|cc|cc}
\hline
Object & \cite{Wang_2021_CVPR} & \cite{Su_2022_CVPR} & \cite{Wang_2021_CVPR}-LC & \cite{Su_2022_CVPR}-LC \\
\hline

002\_master\_chef\_can & 41.5 & \textbf{62.6} & 38.7 & 51.6 \\
003\_cracker\_box & 83.2 & 98.5 & 96.2 & \textbf{99.7} \\
004\_sugar\_box & 91.5 & 96.3 & 98.1 & \textbf{99.4} \\
005\_tomato\_soup\_can & 65.9 &  \textbf{80.5} & 77.6 & 79.6 \\
006\_mustard\_bottle & 90.2 & \textbf{100} & 77.0 & 99.7 \\
007\_tuna\_fish\_can & 44.2 & 70.5 & 63.2 & \textbf{86.1} \\
008\_pudding\_box & 2.8 & \textbf{99.5} & 81.3 & 99.1 \\
009\_gelatin\_box & 61.7 & \textbf{97.2} & 81.8 & 94.9 \\
010\_potted\_meat\_can & 64.9 & \textbf{76.9} & 68.1 & 73.9 \\
011\_banana & 64.1 & 71.2 & 71.0 & \textbf{95.8} \\
019\_pitcher\_base & 99.0 & \textbf{100} & \textbf{100} & \textbf{100} \\
021\_bleach\_cleanser & 73.8 & 75.9 & 69.9 & \textbf{85.6} \\
024\_bowl* & 37.7 & 18.5 & \textbf{44.1} & 35.2 \\
025\_mug & 61.5 & 77.5 & 46.2 & \textbf{88.7} \\
035\_power\_drill & 78.5 & 97.4 & \textbf{99.7} & 99.2 \\
036\_wood\_block* & 59.5 & 87.6 & \textbf{91.7} & 82.6 \\
037\_scissors & 3.9 & \textbf{71.8} & 14.9 & 56.9 \\
040\_large\_marker & 7.4 & 23.3 & \textbf{29.3} & 27.8 \\
051\_large\_clamp* & 69.8 & \textbf{87.6} & 80.5 & 84.4 \\
052\_extra\_large\_clamp* & 90.0 & 98.0 & 95.5 & \textbf{99.1} \\
061\_foam\_brick* & 71.9 & \textbf{99.3} & 57.6 & 91.3 \\
\hline
mean & 60.1 & 80.5 & 70.6 & \textbf{82.4} \\

\hline
\end{tabular}
}
\end{center}
\caption{\textbf{Detailed ADD(-S) scores on YCB-V.} We report the scores of the original baseline methods, GDR-Net~\cite{Wang_2021_CVPR} and ZebraPose~\cite{Su_2022_CVPR}, and also the scores after applying our LC loss, respectively (denoted by ``-LC''). (*) denotes symmetric objects on which the ADD-S score is reported.}
\label{tab:ycbv-detail-add}
\end{table} 
\begin{table*}
\begin{center}
\begin{tabular}{c|cccccc|cc}
\hline
Method & \makecell{RePOSE\\ \cite{Iwase_2021_ICCV}} & \makecell{RNNPose\\ \cite{Xu_2022_CVPR}} & \makecell{SO-Pose\\ \cite{Di_2021_ICCV}} & \makecell{DProST\\ \cite{dprost_2022_eccv}} & \makecell{GDR-Net\\ \cite{Wang_2021_CVPR}} & \makecell{ZebraPose\\ \cite{Su_2022_CVPR}} & GDR-LC & Zebra-LC\\
\hline

ape & 31.1 & 37.18 & 48.4 & 51.4 & 46.8 & 57.9 & 44.44 & \textbf{61.57}\\
can & 80.0 & 88.07 & 85.8 & 78.7 & 90.8 & 95.0 & 89.06 & \textbf{97.35}\\
cat & 25.6 & 29.15 & 32.7 & 48.1 & 40.5 & 60.6 & 49.87 & \textbf{64.49}\\
driller & 73.1 & 88.14 & 77.4 & 77.4 & 82.6 & \textbf{94.8} & 87.81 & 94.65\\
duck & 43.0 & 49.17 & 48.9 & 45.4 & 46.9 & 64.5 & 56.08 & \textbf{66.82}\\
eggbox* & 51.7 & 66.98 & 52.4 & 55.3 & 54.2 & 70.9 & 62.81 & \textbf{71.77}\\
glue* & 54.3 & 63.79 & 78.3 & 76.9 & 75.8 & \textbf{88.7} & 68.88 & 86.35\\
holepuncher & 53.6 & 62.76 & 75.3 & 67.4 & 60.1 & \textbf{83.0} & 72.89 & 81.49\\
\hline
mean & 51.6 & 60.65 & 62.3 & 62.6 & 62.2 & 76.9 & 66.48 & \textbf{78.06}\\

\hline
\end{tabular}
\end{center}
\caption{\textbf{Comparison with the state of the art on LM-O.} (*) denotes symmetric objects on which the ADD-S score is reported. ``GDR-LC'' denotes the LC loss with the GDR-Net~\cite{Wang_2021_CVPR} baseline, ``Zebra-LC'' denotes the LC loss with the ZebraPose~\cite{Su_2022_CVPR} baseline.}
\label{tab:lmo-detail}
\end{table*}
\begin{table*}
\begin{center}
\setlength\tabcolsep{5.4pt}
\begin{tabular}{l|cc|cc|cc|cc}

\hline
 Method & \multicolumn{2}{c|}{GDR-Net~\cite{Wang_2021_CVPR}} &  \multicolumn{2}{c|}{ZebraPose~\cite{Su_2022_CVPR}} &  \multicolumn{2}{c|}{GDR-Net-LC} & \multicolumn{2}{c}{ZebraPose-LC} \\
 \hline
 Metric & \makecell{AUC of \\ADD-S} & \makecell{AUC of \\ADD(-S)} & \makecell{AUC of \\ADD-S} & \makecell{AUC of \\ADD(-S)} & \makecell{AUC of \\ADD-S} & \makecell{AUC of \\ADD(-S)} & \makecell{AUC of \\ADD-S} & \makecell{AUC of \\ADD(-S)} \\
 \hline

002\_master\_chef\_can & *96.3 & *65.2 & 93.7 & 75.4 & 85.6, *90.1 & 57.5, *61.6 & 88.4 & 66.9 \\
003\_cracker\_box & *97.0 & *88.8 & 93.0 & 87.8 & 93.1, *98.1 & 86.8, *91.6 & 93.7 & 88.3 \\
004\_sugar\_box & *98.9 & *95.0 & 95.1 & 90.9 & 95.9, *99.8 & 92.3, *97.4 & 94.7 & 90.3 \\
005\_tomato\_soup\_can & *96.5 & *91.9 & 94.4 & 90.1 & 92.8, *96.2 & 88.2, *93.0 & 93.4 & 89.2 \\
006\_mustard\_bottle & *100 & *92.8 & 96.0 & 92.6 & 94.1, *97.6 & 88.2, *93.1 & 95.1 & 90.9 \\
007\_tuna\_fish\_can & *99.4 & *94.2 & 96.9 & 92.6 & 96.2, *99.9 & 92.1, *96.9 & 97.2 & 94.1 \\
008\_pudding\_box & *64.6 & *44.7 & 97.2 & 95.3 & 94.4, *99.1 & 90.4, *95.3 & 96.7 & 94.7 \\
009\_gelatin\_box & *97.1 & *92.5 & 96.8 & 94.8 & 95.1, *99.9 & 91.7, *96.8 & 96.7 & 94.6 \\
010\_potted\_meat\_can & *86.0 & *80.2 & 91.7 & 83.6 & 85.8, *89.0 & 79.6, *83.8 & 91.3 & 82.5 \\
011\_banana & *96.3 & *85.8 & 92.6 & 84.6 & 92.2, *97.6 & 83.2, *88.0 & 95.3 & 90.1 \\
019\_pitcher\_base & *99.9 & *98.5 & 96.4 & 93.4 & 96.6, *100 & 93.5, *98.4 & 96.4 & 93.2 \\
021\_bleach\_cleanser & *94.2 & *84.3 & 89.5 & 80.0 & 86.3, *91.2 & 77.0, *82.0 & 90.5 & 82.3 \\
024\_bowl* & *85.7 & *85.7 & 37.1 & 37.1 & 83.1, *88.6 & 83.1, *88.6 & 63.9 & 63.9 \\
025\_mug & *99.6 & *94.0 & 96.1 & 90.8 & 92.7, *96.5 & 83.9, *88.9 & 96.5 & 92.3 \\
035\_power\_drill & *97.5 & *90.1 & 95.0 & 89.7 & 96.1, *99.9 & 92.6, *97.9 & 95.4 & 90.8 \\
036\_wood\_block* & *82.5 & *82.5 & 84.5 & 84.5 & 87.1, *92.2 & 87.1, *92.2 & 81.2 & 81.2 \\
037\_scissors & *63.8 & *49.5 & 92.5 & 84.5 & 75.8, *80.4 & 63.5, *67.8 & 88.3 & 79.0 \\
040\_large\_marker & *88.0 & *76.1 & 80.4 & 69.5 & 77.5, *81.8 & 68.8, *73.5 & 77.6 & 68.5 \\
051\_large\_clamp* & *89.3 & *89.3 & 85.6 & 85.6 & 83.1, *87.9 & 83.1, *87.9 & 86.8 & 86.8 \\
052\_extra\_large\_clamp* & *93.5 & *93.5 & 92.5 & 92.5 & 91.4, *95.8 & 91.4, *95.8 & 94.6 & 94.6 \\
061\_foam\_brick* & *96.9 & *96.9 & 95.3 & 95.3 & 90.0, *94.6 & 90.0, *94.6 & 93.2 & 93.2 \\
\hline
mean & *91.6 & *84.4 & 90.1 & 85.3 & 89.8, *94.1 & 84.0, *88.8 & \textbf{90.8} & \textbf{86.1} \\

\hline
\end{tabular}
\end{center}
\caption{\textbf{Detailed AUC scores on YCB-V.}  We report the scores of the original baseline methods and the scores after the LC loss is applied. \mbox{``GDR-Net-LC''} denotes the LC loss with the GDR-Net~\cite{Wang_2021_CVPR} baseline, ``ZebraPose-LC'' denotes the LC loss with the ZebraPose~\cite{Su_2022_CVPR} baseline. A (*) after the object name denotes the symmetric objects on which the ADD-S score is reported. A (*) before the AUC score indicates that the AUC is computed with 11-points interpolation.}
\label{tab:ycbv-detail-auc}
\end{table*} 

{\small
\bibliographystyle{ieee_fullname}
\bibliography{egbib}
}

\end{document}